\pdfoutput=1
\documentclass[a4paper]{article}

\usepackage{INTERSPEECH_v2}
\usepackage{color}
\usepackage{subcaption}
\usepackage{amsmath}
\usepackage{algorithm}
\usepackage{algorithmic}
\usepackage{url}

\title{Deep Reinforcement Learning for Inquiry Dialog Policies \\ with Logical Formula Embeddings}
\name{Takuya Hiraoka$^1$, Masaaki Tsuchida$^1$, Yotaro Watanabe$^2$}
\address{
  $^1$NEC Data Science Research Laboratories, Japan\\
  $^2$PKSHA Technology Inc., Japan}
\email{t-hiraoka@ce.jp.nec.com, m-tsuchida@cq.jp.nec.com, y\_watanabe@pkshatech.com}

\begin{document}

\maketitle
\begin{abstract}
This paper is the first attempt to learn the policy of an inquiry dialog system (IDS) by using deep reinforcement learning (DRL). 
Most IDS frameworks represent dialog states and dialog acts with logical formulae. 
In order to make learning inquiry dialog policies more effective, 
we introduce a logical formula embedding framework based on a recursive neural network. 
The results of experiments to evaluate the effect of 1) the DRL and 2) the logical formula embedding framework show that the combination of the two are as effective or even better than existing rule-based methods for inquiry dialog policies. 
\end{abstract}
\noindent\textbf{Index Terms}: inquiry dialog, reinforcement learning, logical formula embedding

\section{Introduction}\label{sec:intro} 
The objective of inquiry dialogs is to cooperatively answer questions (or problems) shared by participants~\cite{walton1995commitment}. 
In inquiry dialogs, participants  (including the dialog system) do not have complete domain knowledge, so they share their own knowledge with their partners. 
This setting is different from slot-filling dialog settings (e.g., \cite{williams2007partially,levin2000stochastic,walker2000application}), where the systems are required to have complete domain knowledge. 
Thus, the realization of practical IDSs extends the capabilities of current dialog systems. 
In addition, IDSs can actively expand their own knowledge bases through dialogs, which helps to reduce the costs of manual knowledge base expansion. 

Although there has been previous research on IDS, the focus has not been on learning its policies. 
Amgoud, Parsons and McBurney have discussed the fundamental principles of inquiry dialog~\cite{amgoud2000modelling,parsons2002analysis,parsons2003outcomes,mcburney2001representing}, 
and Black and Hunter proposed policies for inquiry dialogs~\cite{black2007generative,black2009inquiry}. 
In addition, Fan and Toni proposed policies for inquiry and information-seeking dialogs~\cite{fan2012agent,fan2015mechanism}. 
These studies dealt with rule-based policies that only work efficiently in limited situations; learning efficient policies in various conditions remains still an open problem. 

In this work, we apply DRL to learn the policies for IDS. 
In addition, in order to make learning these policies more effective, we propose a logical formula embedding framework. 
In Section \ref{sec:inq}, we introduce our inquiry dialog domain and IDS framework. 
We show how dialog states and dialog acts are represented by logical formulae in the framework. 
In Section \ref{sec:method}, we explain how to apply DRL to inquiry dialogs. 
Specifically, we use ``Deep Q-Learning with experience replay'' (DQL)~\cite{mnih2013playing,mnih2015human} for DRL and introduce a logical formula embedding framework to it. 
In Section \ref{sec:eval}, we evaluate the effectiveness of the DRL and the logical formula embedding framework for producing good policies. 
We conclude the paper in Section \ref{sec:conc} with a brief summary. 

Our research contribution is two-fold: 
first, to our knowledge, this is the first study that applies reinforcement learning to learning inquiry dialog policies, 
and second, we introduce logical formula embedding frameworks for DRL in order to make learning inquiry dialog policies more effective.

\section{Inquiry Dialog}\label{sec:inq}
\subsection{Inquiry Dialog and its Domain}
In inquiry dialogs, an IDS and its user collaborate in order to answer their shared questions. 
They start the dialog with shared questions (\textbf{queries}), 
and then, in order to come up with an answer, they share what they believe in (\textbf{beliefs}) by reasonable assertion (\textbf{arguments}) and then make new arguments reflecting those shared beliefs. 
Sharing their beliefs is continued until they find an answer (or find that there is no possible answer). 
The shared beliefs are stored in the \textbf{commitment store}, and participant's beliefs are stored in the \textbf{belief base}.

As an example of inquiry dialog domains, we discuss a system and its user working on \textit{Compliance Violation Detection}. 
In this setting, the system and the user play detectives. 
They are given different information sources from which they extract information, and then start discussing to answer the query ``Is there a compliance violation?". 
An example of one participant's beliefs and a dialog is shown in Table \ref{tab:belexample}. 
The beliefs pertain to information (such as the e-mail contents of suspects) that they surveyed and to compliance violation. 
An example of beliefs is shown in Table \ref{tab:belexample}a. 
The belief  ``Company A proposed a price to another company" is contained in the user's belief base, 
and the beliefs ``Company B accepted a proposal" and ``If company A proposed a price and company B accepted it, it is a compliance violation" are  contained in the system's belief base. 
They exchange their arguments in order to answer the given query. 
Examples of arguments and how to share beliefs are shown in Table \ref{tab:belexample}b. 
The user conveys his belief ``Company A proposed a price in the email." to the system, with the argument ``Company A proposed a price  and company B accepted it." 
After that, the system argues ``Company B accepted a proposal. If the proposal is made by company A, it is a compliance violation." reflecting the belief shared by the user. 
The system's argument is the answer to the query. 
After that, both participants express that they have nothing to discuss anymore, and the dialog is closed. 
\begin{table*}[!th]
  \begin{center}
    \caption{An example of an inquiry dialog for \textit{Compliance Violation Detection}. The upper part of the table (Table \ref{tab:belexample}a) lists the participant's beliefs and the given query, and the lower part (Table \ref{tab:belexample}b) gives an example dialog. }
    \vspace{-10pt}
    \label{tab:belexample}

    \scalebox{0.8}{\begin{tabular}{|l|l|l|} \hline
    & Transcription of belief & Logical formula representation  \\ \hline \hline

    System's belief base $\Sigma_{Sys}$ & Company B accepted a proposal. & $\mbox{CompanyB}(z) \wedge \mbox{Accept}(e2, z, w)$ \\ \cline{2-3}
    & If company A proposed a price and company B accepted it, & $\mbox{CompanyA}(X) \wedge \mbox{Propose}(E1, X, Y) \wedge \mbox{CompanyB}(Y) $ \\
    & it is a compliance violation. & $\wedge \mbox{Accept}(E2, Y, E1) \rightarrow \mbox{ComplianceViolation}(E3, X, Y)$ \\ \hline

    User's belief base $\Sigma_{Usr}$ & Company A proposed a price to another company.  & $\mbox{CompanyA}(x) \wedge \mbox{Propose}(e1, x, y)$ \\ \hline

    Query & Is there a compliance violation? & $ \rightarrow \mbox{ComplianceViolation}(E3, X, Y)$ \\ \hline
    \end{tabular}}

\vspace{+5pt}\scalebox{0.75}{\begin{tabular}{|l||l|l|l|l|l|} \hline
      i  & sp & Natural language transcription / dialog act & Commitment store CS & \multicolumn{2}{l|}{Current focus of query store cQS} \\ \hline \hline
      0 &  & $\begin{pmatrix} \mbox{Dialog is started with the query} \ \mbox{``Is there a compliance violation?"} \end{pmatrix} $  &  & \multicolumn{2}{l|}{$\mbox{ComplianceViolation}(E3, X, Y)$} \\ \cline{1-3}\cline{6-6}
      1 & Sys. &  If company A proposed a price and company B accepted it, &  &&  $\mbox{CompanyA}(X), \mbox{Propose}(E1, X, Y), $ \\ 
        &        & it is a compliance violation. Let's discuss it. &  && $\mbox{CompanyB}(Y), \mbox{Accept}(E2, Y, E1), $ \\ \cline{3-3}
        &        & Open$\begin{pmatrix} \mbox{CompanyA}(X) \wedge \mbox{propose}(E1, X, Y) \wedge \mbox{CompanyB}(Y) \\ \wedge \mbox{Accept}(E2, Y, E1) \rightarrow \mbox{ComplianceViolation}(E3, X, Y) \end{pmatrix}$ & && $\mbox{ComplianceViolation}(E3, X, Y)$ \\ \cline{1-3}
      2 & Usr. & Company A proposed a price in the e-mail.  & &&  \\ \cline{3-3}
         &        & Assert$\begin{pmatrix} \{\mbox{CompanyA}(x) \wedge \mbox{Propose}(e1, x, y)\},\\ \mbox{CompanyA}(x) \wedge \mbox{Propose}(e1, x, y)\end{pmatrix}$ & 
           $\begin{matrix} \mbox{``Company A proposed a price to another company."} \\ \{\mbox{CompanyA}(x) \wedge \mbox{Propose}(e1, x, y)\}  \end{matrix}$ &&  \\ \cline{1-3}

      3 & Sys. & Company B accepted a proposal from another company.  & &&  \\ 
        &        & If the proposal is made by company A, it is a compliance violation. & &&  \\ \cline{3-3}
        &        & Assert$\begin{pmatrix} \begin{Bmatrix} 
  \mbox{CompanyA}(x) \wedge \mbox{Propose}(e1, x, y), \\
  \mbox{CompanyB}(z) \wedge \mbox{Accept}(e2, z, w), \\
  \mbox{CompanyA}(X) \wedge \mbox{Propose}(E1, X, Y) \\ \wedge \mbox{CompanyB}(Y) \wedge \mbox{Accept}(E2, Y, E1) \\ 
  \rightarrow \mbox{ComplianceViolation}(E3, X, Y)
  \end{Bmatrix}, \\ \mbox{ComplianceViolation}(e3, x, z) \end{pmatrix}$ &
    $\begin{matrix} 
      \mbox{``Company B accepted a proposal."}\\
      \{\mbox{CompanyB}(z) \wedge \mbox{Accept}(e2, z, w)\}\\ \\

      \mbox{`` If company A proposed a price}\\
      \mbox{and company B accepted it, }\\
      \mbox{it is a compliance violation."}\\
      \begin{Bmatrix} \mbox{CompanyA}(X) \wedge \mbox{Propose}(E1, X, Y) \\ \wedge \mbox{CompanyB}(Y) \wedge \mbox{Accept}(E2, Y, E1) \\ \rightarrow \mbox{ComplianceViolation}(E3, X, Y) \end{Bmatrix}
    \end{matrix}$
   &&  \\ \cline{1-3}
      4 & Usr. & True. I don't have anything to discuss on the current topic. / Close() &  &&  \\ \cline{1-3}
      5 & Sys. & Me neither. / Close() &  &&  \\ \cline{1-3}\cline{6-6}
      6 & Usr. & Do you still have anything to say about the topic  &&\multicolumn{2}{l|}{}  \\ 
       & & "Is there a compliance violation?"  &&\multicolumn{2}{l|}{} \\ 
       & & I don't have anything further to say. / Close() &&\multicolumn{2}{l|}{}  \\ \cline{1-3}
      7 & Sys. & No.  / Close() &&\multicolumn{2}{l|}{}  \\ \hline
      \multicolumn{6}{|c|}{$\begin{pmatrix} \mbox{Dialog is closed} \end{pmatrix}$}  \\ \hline
    \end{tabular}}
  \end{center}
\vspace{-26pt}
\end{table*}

\subsection{Framework for Inquiry Dialog Systems}\label{sec:framework}
Our framework basically follows the IDS framework Black and Hunter proposed~\cite{black2007generative,black2009inquiry} because it is clearly defined enough to be implemented easily. 

In our framework, beliefs are represented with (first-order predicate) logical formulae. 
In first-order predicate logic, a logical formula is composed of an atom $a(X,..)$, a ground atom $a(x,..)$, or their negations $\neg a(X,..), \neg a(x,..)$. 
Beliefs are categorized into two types according to the form of the logical formula: 
\vspace{-5pt}
\begin{description}
\setlength{\itemsep}{0mm} 
\setlength{\parskip}{0mm} 
\item[State belief]:  represented by conjunction of ground atoms: $a_{1}(x_{1},...) \wedge ... \wedge a_{n}(x^{'}_{1},...)$, 
\item[Domain belief]: represented by an inference rule: \\ $a_{1}(X_{1},...) \wedge ... \wedge a_{n}(X^{'}_{1},...) \rightarrow a_{0}(X^{''}_{1},...) $, 
\end{description}
\vspace{-5pt}
where $\wedge$ represents logical conjunction and $\rightarrow $ represents implication. 
The set of all possible beliefs is denoted by $\mathcal{B}$. 
The third column of Table \ref{tab:belexample}a shows examples of beliefs. 
Representing beliefs with logical formulae is commonly seen in most of previous research on IDSs. 

Dialog states are composed of 1) system beliefs, 2) a commitment store, and 3) a query store. 
System belief $\Sigma_{Sys.}$ and commitment store $CS$ are subsets of $\mathcal{B}$. 

Dialog acts are \textbf{Assert}, \textbf{Open}, and \textbf{Close}: 
\vspace{-5pt}
\begin{description}
\setlength{\itemsep}{0mm} 
\setlength{\parskip}{0mm} 
\item[Assert($\Phi, \phi$)]: represents an asserting argument ($\Phi, \phi$). 
$\phi$ is a domain belief, called a \textbf{claim}. 
$\Phi$ is a set of beliefs, called \textbf{supports} of the claim. 
$\Phi$ should be a minimal and consistent belief set for deriving the claim $\phi$. 

\item[Open($\Omega$)]: represents the intention to start a discussion on agenda $\in \Omega$. $\Omega$ is a domain belief. 

\item[Close()]: represents the intention to close the discussion on the current agenda. 

\end{description}
\vspace{-5pt}
In the dialog example in Table \ref{tab:belexample}b, Assert is shown at $i=2, 3$,  Open is shown at $i=1$, and Close is shown at $i=4, 5, 6, 7$. 

The goal of the dialog is to generate arguments to answer the query. 
More concretely, given a query $ \rightarrow q_{1}(X_{1},...) \wedge ... \wedge q_{n}(X^{'}_{1},...)$, the dialog succeeds if either of the participants performs $\mbox{Assert}(\Phi^*, \phi^*)$, s.t. $\phi^{*}= q_{1}(x_{1},...) \wedge ... \wedge q_{n}(x^{'}_{1},...)$. 
In the example given in Table \ref{tab:belexample}b, the query is given as ``$\rightarrow \mbox{ComplianceViolation}(E3, X, Y)$." 
The Assert at $i=3$ addresses the query, and thus achieves the goal of the dialog. 

If a participant performs Assert($\Phi, \phi$) at time point $i$, 
the commitment store $CS$ is updated ($CS_{i} \leftarrow CS_{i-1} \cup \Phi$). 
In the example of Table \ref{tab:belexample}b, $CS$ is updated by an Assert at $i=2, 3$. 

We introduce a stack of \textbf{query stores} $cQS$ in order to manage the current agenda of a dialog. 
A query store is a list of atoms that represent the current agenda of the dialog. 
$cQS$ is a stack of query stores and it is updated when participants perform Open or Close. 
If either of the participants perform Open, a new query store is stacked to $cQS$. 
In addition, if all participants perform Close, the top of $cQS$ is removed. 
$cQS$ is initialized with the query. 
In Table \ref{tab:belexample}b, $cQS$ is initialized with the query, and ``$\mbox{ComplianceViolation}(E3, X, Y)$" is stacked to $cQS$. 
In addition, a query store of five atoms is stacked to $cQS$ by Open at $i=1$ and is removed when both the system and the user perform Close ($i=4, 5$). 

The set of valid dialog acts in the given dialog state is called \textbf{legal moves}. 
In inquiry dialog, Assert and Open should be related to the current agenda. 
More concretely, at time point i, the legal moves of Assert $L_{assert, i}$, Open $L_{open, i}$, Close $L_{close, i}$, for the system are defined as follows: 
\vspace{-5pt}
\begin{description}
\setlength{\itemsep}{0mm} 
\setlength{\parskip}{0mm} 
\item[$L_{assert,i}$] $\{ Assert(\Phi, \phi) |$ \\
                                                            1) $\phi$ is a ground atom corresponding to an atom in the list at $cQS$ top, \\ 
                                                            2) $\Phi \subseteq (\Sigma_{Sys} \cup CS_{i}) \}$
\item[$L_{open,i}$] $\{ Open(a_{1}(X_{1},...) \wedge ... \wedge a_{n}(X^{'}_{1},...) \rightarrow a_{0}(X^{''}_{1},...)) |$  \\ 
                                                            1) $a_{0}(X^{''}_{1},...) \mbox{is an element of the top of } cQS$, \\
                                                            2) $a_{1}(X_{1},...) \wedge ... \wedge a_{n}(X^{'}_{1},...) \rightarrow a_{0}(X^{''}_{1},...) \in \Sigma_{Sys}\}$ 
\item[$L_{close,i}$] $\{ Close() \}$
\end{description}
\vspace{-5pt}
The system can not utilize an Assert or Open that has already been performed in the past.

\section{Methodology for Learning Inquiry Dialog Policies}\label{sec:method}

\subsection{Learner's Model}\label{sec:learnmodel}
We define rewards, actions, and states in Markov decision processes (MDPs)~\cite{sutton1998reinforcement} to apply DRL for learning IDS policies. 

A reward is structured in order for the system to answer the given queries as fast as possible. 
The reward $r_t$ at each system's turn $t$ is fed with the following equation: 
\vspace{-5pt}
\begin{equation}
  r_t = \begin{cases}
    w_{pos} & (\mbox{if either of the participants answer the query}) \\
    -w_{neg} & (\mbox{otherwise})
  \end{cases}, 
\label{rew:orig} \nonumber
\vspace{-5pt}
\end{equation}
where $w_{pos}$ and $w_{neg}$ are numerical values $[0, \infty)$. 
The positive reward $w_{pos}$ is fed if either the system or its user assert the answer to the given question. 
In addition, the negative reward $-w_{neg}$ representing time pressure is fed at each turn's end\footnote{Namely, a time point immediately after the system updates its dialog state with a user's dialog act. }.  
We set $w_{pos}$ to 20 and $w_{neg}$ to 1. 

Action is represented by a dialog act from the legal moves, and the dialog state represents the state (described in Section \ref{sec:framework}). 
There are multiple ways to encode dialog state/act to state/action (in MDPs). 
A naive one is the \textit{bag of logical formulae}, namely, state and action are represented by binary vectors whose element is set to 1 if corresponding logical formula is contained in a dialog act or dialog state. 
However, the problem with this representation is that state/action space easily becomes very sparse because the size of the vectors is determined by that of all possible beliefs $|\mathcal{B}|$.

\subsection{Logical Formula Embedding Framework}\label{sec:embf}
If we want DRL to efficiently produce better policies, it makes sense to represent dialog states and dialog acts compactly as state and action in MDPs. 
As seen in the Section \ref{sec:framework}, dialog states and dialog acts are represented with beliefs (i.e., logical formulae) in typical inquiry dialog frameworks. 

\begin{figure}[!t]
  \begin{center}
    \includegraphics[width=0.99\hsize]{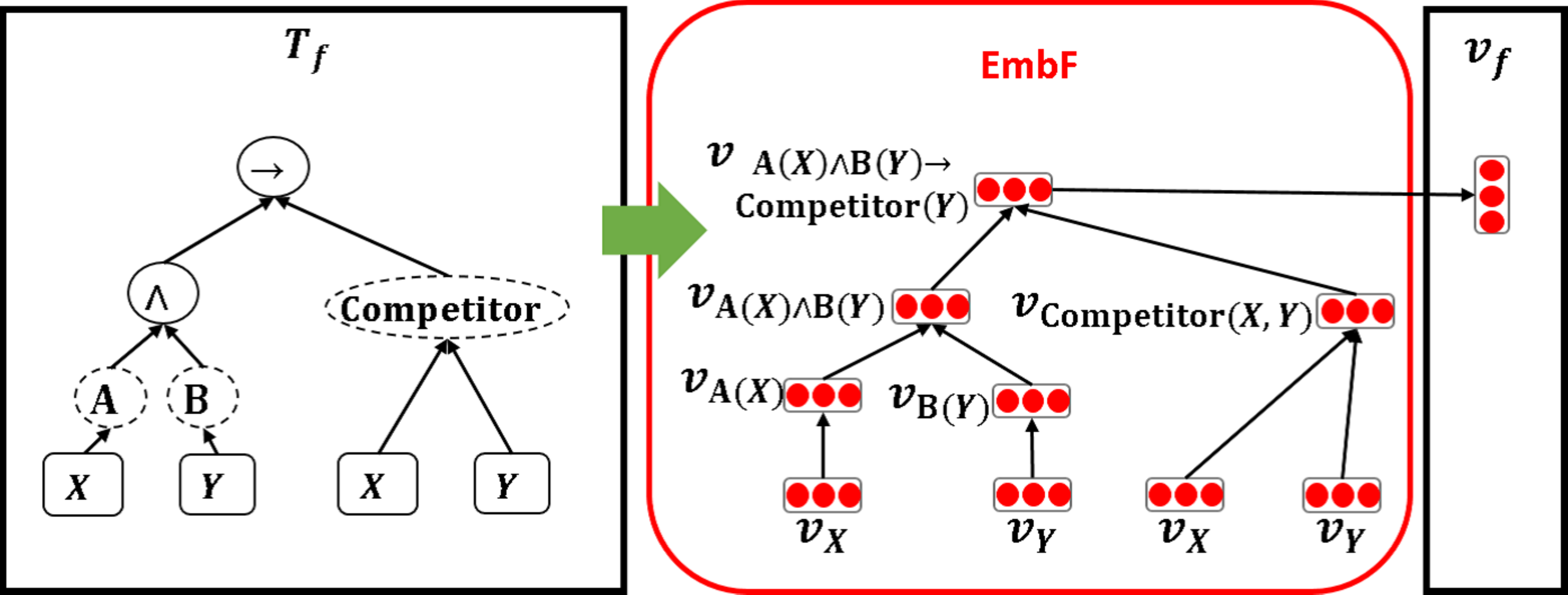}
    \vspace{-8pt}
    \caption{The framework of the logical formula embedding. }
    \label{fig:lfe}
  \end{center}
\vspace{-5pt}
\end{figure}

We propose a recursive neural network~\cite{socher2011parsing,socher2013recursive} based framework for injecting logical formula into embeddings (``EmbF" in Fig.~\ref{fig:lfe}). 
The proposed framework (``EmbF") calculates the compositional vector $v_{f}$ of the tree representation $T_{f}$ of a logical formula $f$ in a bottom up manner. 
$T_{f}$ is an abstract syntax tree of the truth assignment for $f$~\cite{rocktaschel2015injecting,guo2016jointly}. 
The leaf nodes in $T_{f}$ represent arguments of predicates, and internal nodes represent either predicates or logical operators. 
Note that parent nodes of leaf nodes must represent predicates. 
In Fig.~\ref{fig:lfe},  ``$\mbox{A}(X) \wedge \mbox{B}(Y) \rightarrow \mbox{Competitor}(X, Y)$" is parsed into tree representation. 
The leaf nodes represent the arguments ``X and Y", the dashed internal nodes represent the predicates ``A, B and Competitor", and the other internal nodes represent logical operators ``$\wedge \mbox{ and } \rightarrow$." 

Traversing $T_{f}$, EmbF  injects atoms into embeddings and composes them recursively. 
First, EmbF calculates atom (i.e., leaves and their parent nodes) embeddings $v_{a}$ on the basis of 
\vspace{-5pt}
\begin{eqnarray}
  v_{a} & = & \mbox{sigmoid}\begin{pmatrix} W_{pre} v_{pre} + W_{arg} \begin{bmatrix} v_{arg,1} \\ \vdots \\ v_{arg, N} \end{bmatrix} \end{pmatrix}, \nonumber
\vspace{-5pt}
\end{eqnarray}
where $W_{pre} \in \mathbb{R}^{d \times  d_{pre}} $ and $W_{arg} \in \mathbb{R}^{d \times  N d_{arg}} $ represent weight matrices, 
$d$ is the number of dimensions of the embedding vector, $d_{pre}$ is that of $v_{pre}$, $d_{arg}$ corresponds to $v_{arg, 1} \hdots v_{arg, N}$, $v_{pre}$ represents the vector of the predicate, and $v_{arg, 1} \hdots v_{arg, N}$ represent the vectors of the arguments. 
We use one-hot representation for $v_{pre}$ and  $v_{arg, 1} \hdots v_{arg, N}$. 
Next, on the basis of the internal nodes representing logical operators ($\wedge$ or $\rightarrow$) and their child nodes labeled $f_{1}, f_{2}$, they compose children's vectors with 
\vspace{-5pt}
\begin{eqnarray}
  v_{f_{1} \wedge f_{2}} & = & \mbox{sigmoid}\begin{pmatrix} W_{\wedge} \begin{bmatrix} v_{f_{1}} \\ v_{f_{2}} \end{bmatrix} \end{pmatrix},  \nonumber \\
  v_{f_{1} \rightarrow f_{2}} & = & \mbox{sigmoid}\begin{pmatrix} W_{\rightarrow} \begin{bmatrix} v_{f_{1}} \\ v_{f_{2}} \end{bmatrix} \end{pmatrix}, \nonumber
\vspace{-5pt}
\end{eqnarray}
where $W_{\wedge}, W_{\rightarrow} \in \mathbb{R}^{d \times 2d}$ represent weight matrices\footnote{As truth assignments $f_{1} \wedge f_{2}$ and $f_{2} \wedge f_{1}$ are identical, \textcolor{black}{we could use other equations that calculate $v_{f_{1} \wedge f_{2}}$, s.t. $v_{f_{1} \wedge f_{2}} = v_{f_{2} \wedge f_{1}}$. Comparisons with  the case of using these equations are left for future work.} }. 
$v_{f_1}, v_{f_2}$ are vector representations of the children's vector. 
The vector $v_{f}$ of the top node is then used as compact belief representation for encoding the dialog state/act. 

\begin{figure}[!t]
  \includegraphics[width=0.99\hsize]{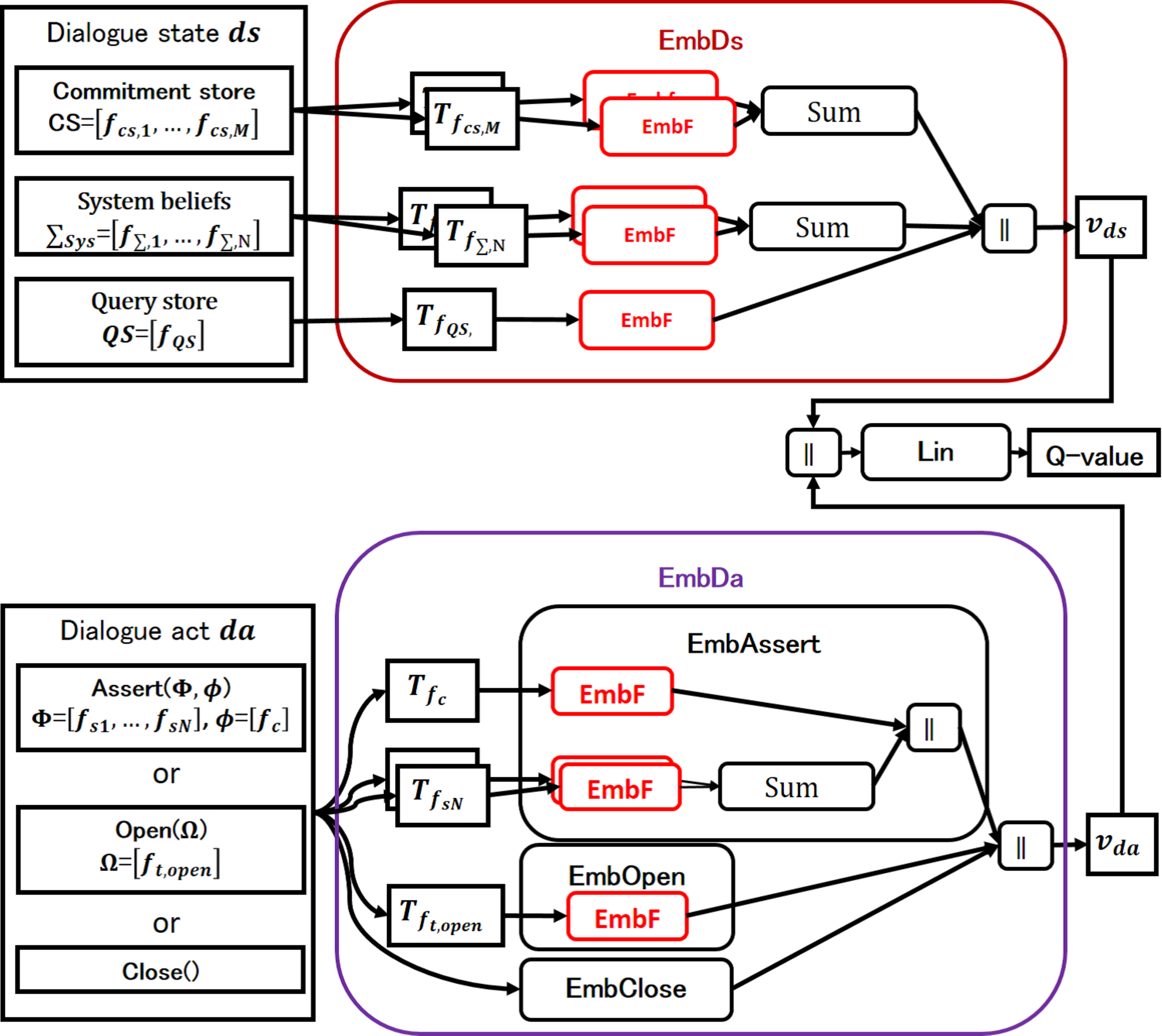}
    \vspace{-8pt}
  \caption{Implementation of the Q-function. Rounded rectangles represent functions, normal rectangles represent data, and arrows represent data flows. ``Lin" represents a linear function.  ``Sum" represents the element-wise sum of vectors. ``$||$" represents the concatenation of vectors. ``EmbClose" is an one hot vector showing the appearance of Close in the dialog act. }\label{fig:qnet}
\vspace{-10pt}
\end{figure}

We implement a Q-function $Q(ds, da)$ that evaluates the expectation of the cumulative future reward (Q-value) of the dialog act and dialog state pair (Fig.~\ref{fig:qnet}). 
$Q(ds, da)$ takes as input a pair consisting of the dialog state $ds$ and the dialog act $da$, and the input is forwarded to the embedding framework for the dialog state (EmbDs) and that for the dialog act (EmbDa). 
EmbDs and EmbDa calculate vector representation $v_{ds}, v_{da}$ of $ds$ and $da$, respectively. 
These vectors are compositions of vectors of logical formula $v_f$  produced by EmbF (discussed above). 
In both EmbDs and EmbDa, the logical formulae in the dialog state and the dialog act are embedded into numerical vectors by EmbF. 
Finally, the Q-function estimates the Q-value with these embedded vectors. 
Parameters of the Q-function (i.e., $W_{\wedge}$, $W_{\rightarrow}$, and weights for ``Lin") are learned with DQL\footnote{Note that our logical formula embedding framework can be applied to other DRLs (e.g., ~\cite{lever2014deterministic,schaul2015prioritized,lillicrap2015continuous,van2016deep}) as well. }. 

\subsection{User Simulator}\label{sec:su}
We utilize user simulators instead of real human users to train the system. 
The user simulator selects a dialog act $da$ in accordance with its policy, which is a hybrid combining a rule-based policy and a random policy. 
With probability $p$, the policy follows the ruled-based one, proposed in previous research~\cite{black2007generative,black2009inquiry} (Algorithm \ref{alg:baseline}). 
The rule-based policy is designed such that the user shares all of his or her beliefs exhaustively with the partner. 
Given legal moves $L_{assert, i}, L_{open, i}, L_{close, i}$, the policy checks if each of the legal moves is empty or not in the order $L_{assert,i}, L_{open,i}, L_{close,i}$. 
If the legal moves are not empty, it selects $da$ from the legal moves. 
In contrast, with the probability $(1-p)$, the random policy is followed. 
The random policy selects $da$ from a set of dialog acts that do not conflict with the user belief base and commitment store. 
\begin{algorithm}[!t]
\caption{Black and Hunter's exhaustive policy}
\label{alg:baseline}
\begin{algorithmic}
\REQUIRE Legal moves $L_{assert, i}, L_{open, i}, L_{close, i}$

\IF{$L_{assert} \neq \emptyset$}
\STATE randomly select a response $da$ from $L_{assert, i}$
\ELSIF{$L_{open} \neq \emptyset$}
\STATE randomly select a response $da$ from $L_{open, i}$
\ELSE
\STATE randomly select a response $da$ from $L_{close, i}$
\ENDIF
\STATE return $da$
\end{algorithmic}
\end{algorithm}\setlength{\textfloatsep}{5pt}

\section{Experimental Evaluation}\label{sec:eval}
In this section, we elucidate the effects of DRL and the logical formula embeddings proposed in Section \ref{sec:embf}. 
Four policies in six different experimental setups are compared. 
These four policies are as follows: 
\vspace{-5pt}
\begin{description}
\setlength{\itemsep}{0mm} 
\setlength{\parskip}{0mm} 
\item[Baseline]: 
The rule-based policy that follows Algorithm \ref{alg:baseline}. 
\item[DQLwoE]: 
The policy produced by the DQL without logical formula embedding proposed in Section \ref{sec:embf}. 
It follows the \textit{bag of logical formulae} (Section \ref{sec:learnmodel}). 
\textcolor{black}{Also, following the previous work on DRL for dialog systems~\cite{cuayahuitl2017simpleds}, it utilizes a multilayer neural net for the Q-function. }
\item[DQLwE-5d]: 
The policy produced by the DQL with logical formula embedding proposed in Section \ref{sec:embf}, with $d=5$. 
\item[DQLwE-10d]: The same as DQLwE-5d except that  $d=10$. 
\end{description}
\vspace{-5pt}
We consider 2000 dialogs as one epoch, and learning is finished when the number of epochs becomes 100 (200,000 dialogs). 
The policy at the end of learning is used in the evaluation. 
We set the discount rate to 0.99, and use an $\epsilon$-greedy policy. $\epsilon$ is linearly annealed from 1.0 to 0.05 during the first 50 epochs. 

The six experimental setups differ in 1) user behavior and 2) the initial condition of the system belief. 
We experiment with two types of user simulator proposed in Section \ref{sec:su}: 
\vspace{-5pt}
\begin{description}
\setlength{\itemsep}{0mm} 
\setlength{\parskip}{0mm} 
\item[RandU]: User simulator with $p=0.75$. 
\item[RuleU]: User simulator with $p=1$. 
\end{description}
\vspace{-5pt}
In addition, we experiment with three different initial conditions of the system belief. 
We prepared a set of 13 beliefs related to the \textit{Compliance Violation  Detection} domain. 
Seven of these beliefs are about the violation of compliance and the remaining six are about the content of the e-mail threads. 
In all experiments, the beliefs about the content of the e-mail threads are randomly assigned to either the system or the user. 
We assign the beliefs of compliance violation in three different ways: 
\vspace{-5pt}
\begin{description}
\setlength{\itemsep}{0mm} 
\setlength{\parskip}{0mm} 
\item[RB]: Each belief is assigned to either the system or the user randomly. 
\item[UB]: All beliefs are assigned to the user. 
\item[SB]: All beliefs are assigned to the system
\end{description}
\vspace{-5pt}
Note that these scenarios vary in the amount of knowledge that the system possesses about compliance violation. 

The performance of the learned policies is evaluated in the success rate at each turn. 
The success rate represents the percentage of dialogs where either the system or the user asserts the answer to the query. 
The success rate is calculated on the basis of 2000 simulated dialogs. 

\begin{figure}[!t]
\begin{minipage}{0.5\hsize}
\begin{center}
\includegraphics[width=1\hsize]{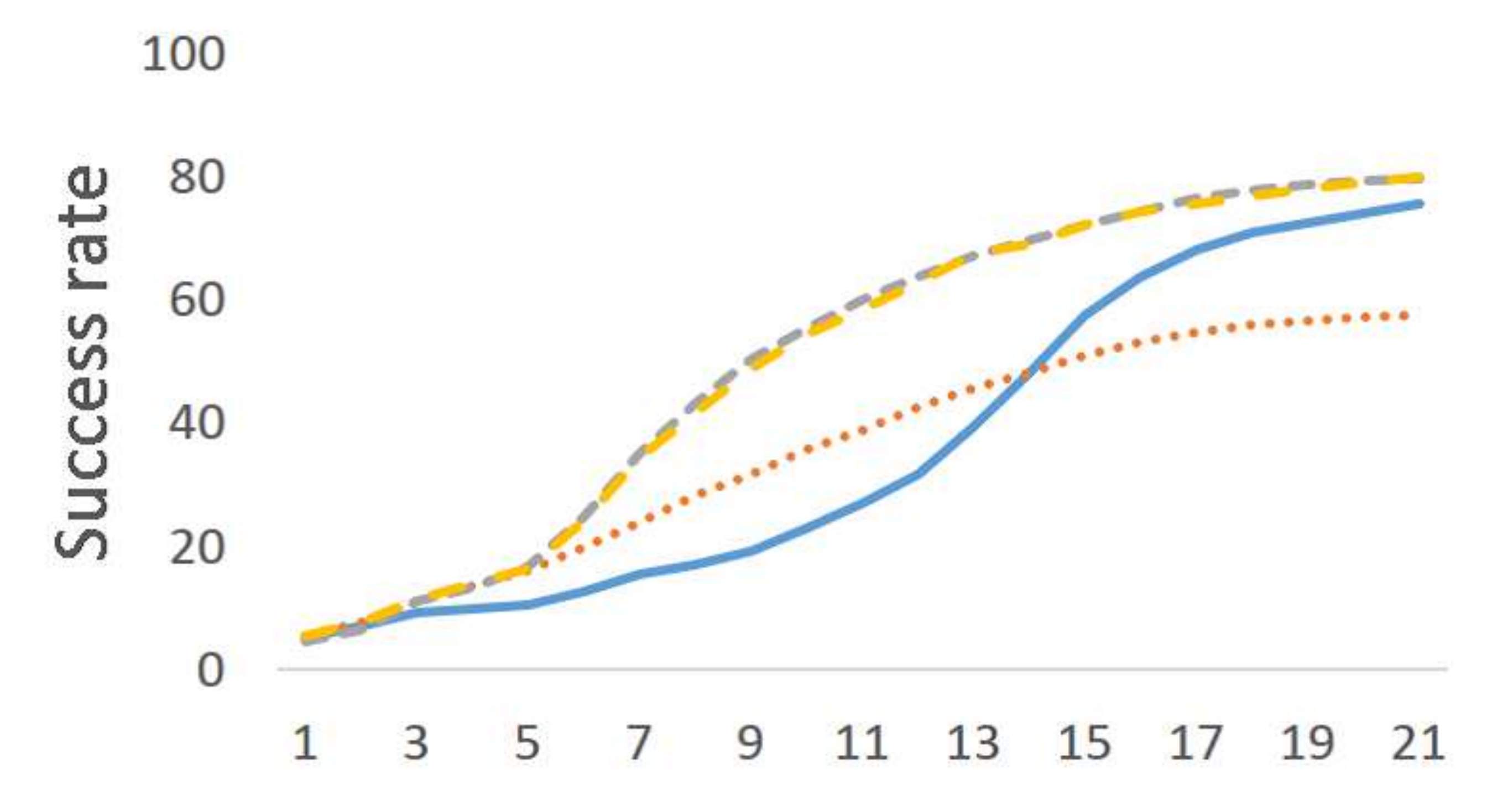}
\end{center}
    \vspace{-14pt}
\subcaption{\textbf{RB}$\wedge$\textbf{RandU}}
\label{fig:11}
\end{minipage}\begin{minipage}{0.5\hsize}
\begin{center}
\includegraphics[width=1\hsize]{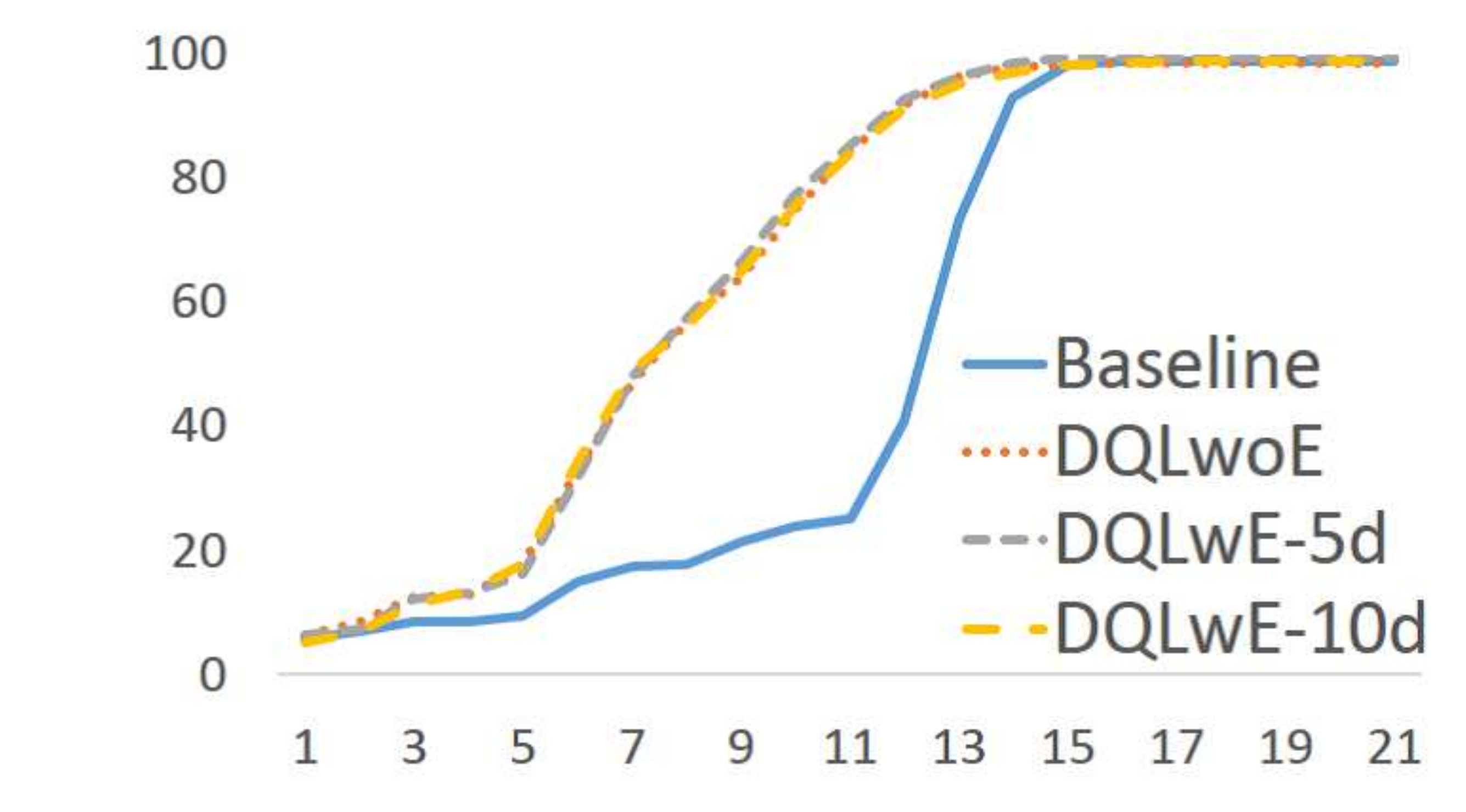}
\end{center}
    \vspace{-14pt}
\subcaption{\textbf{RB}$\wedge$\textbf{RuleU}}
\label{fig:12}
\end{minipage}
\begin{minipage}{0.5\hsize}
\begin{center}
\includegraphics[width=1\hsize]{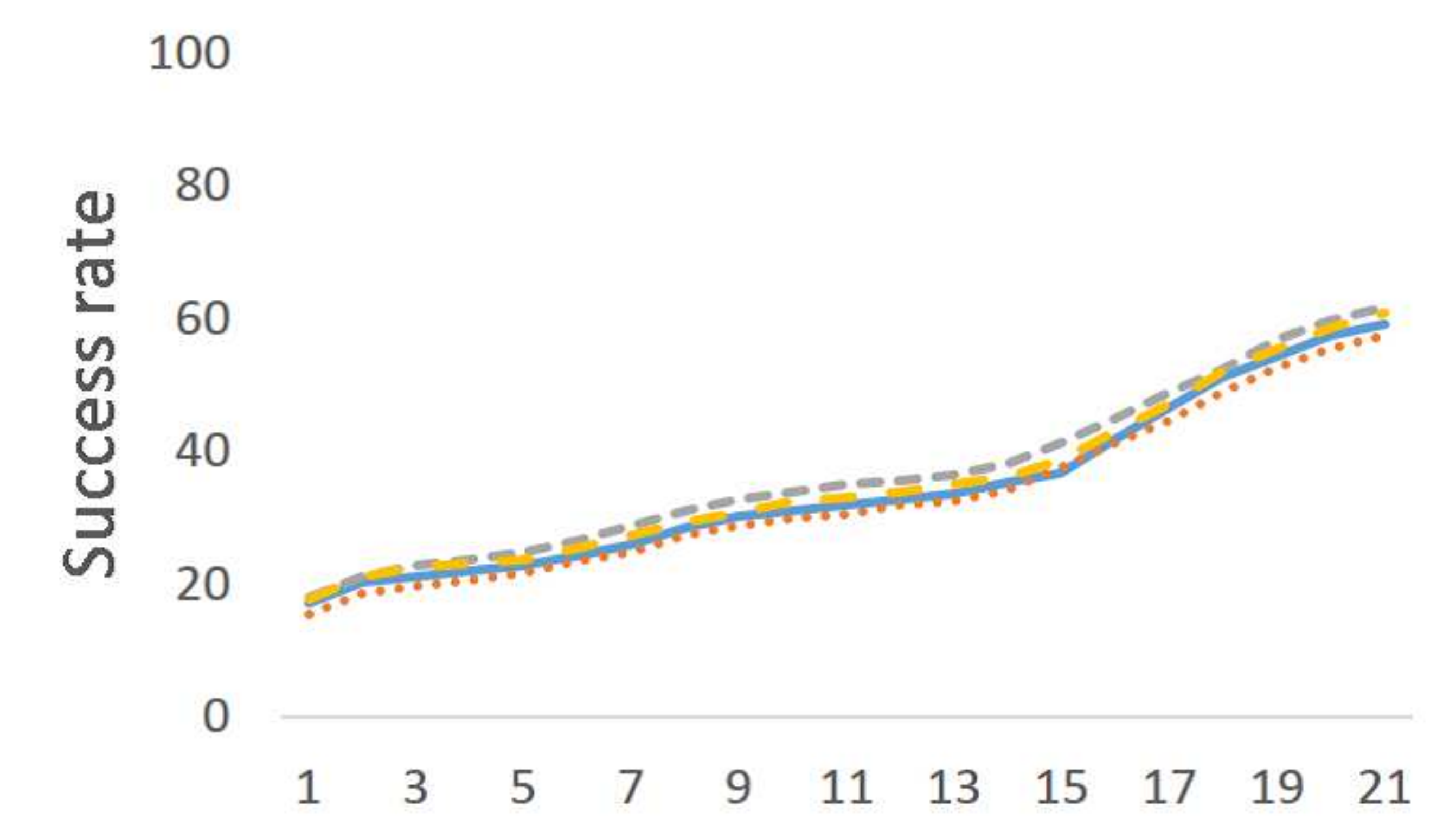}
\end{center}
    \vspace{-14pt}
\subcaption{\textbf{UB}$\wedge$\textbf{RandU}}
\label{fig:21}
\end{minipage}\begin{minipage}{0.5\hsize}
\begin{center}
\includegraphics[width=1\hsize]{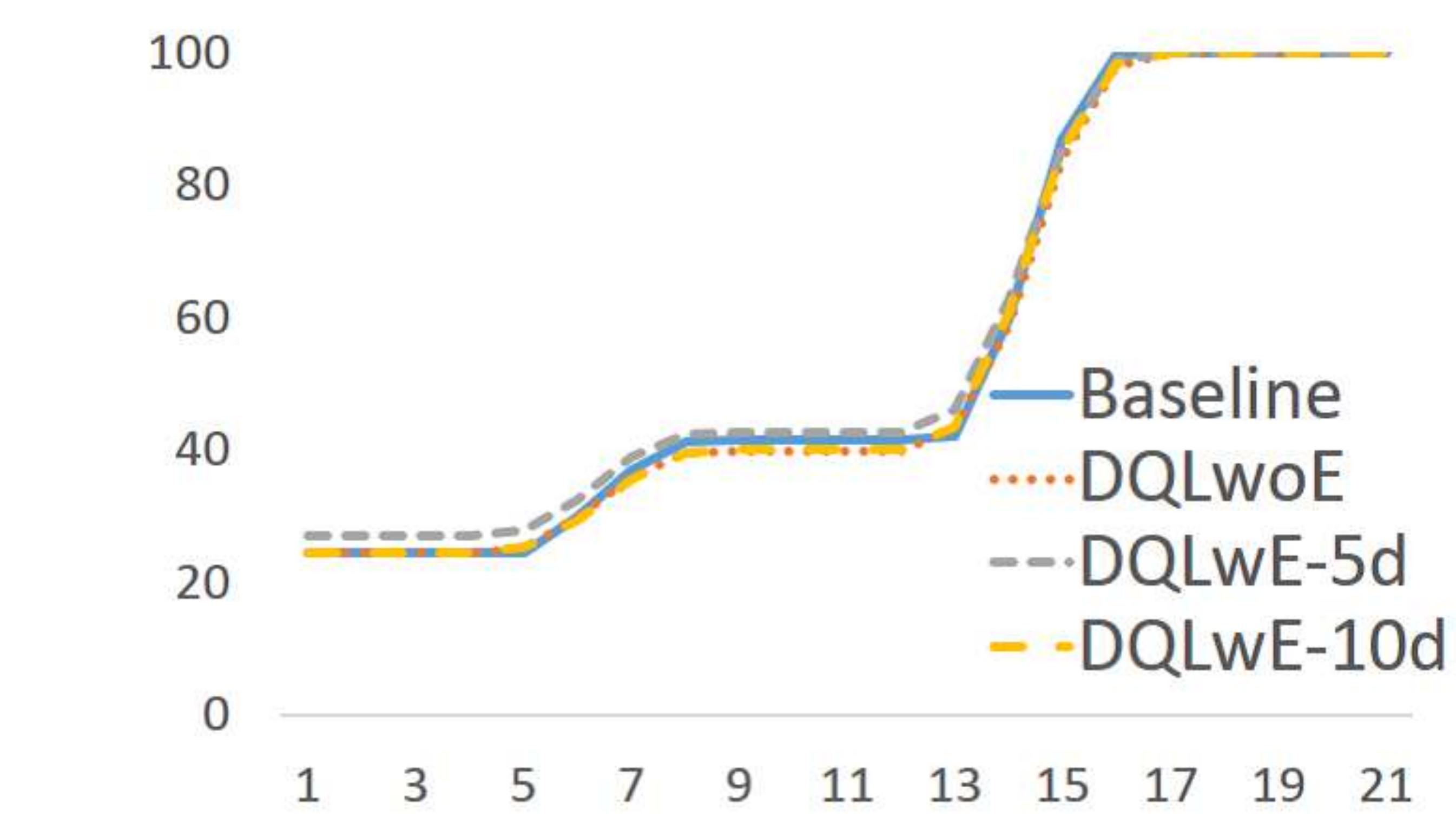}
\end{center}
    \vspace{-14pt}
\subcaption{\textbf{UB}$\wedge$\textbf{RuleU}}
\label{fig:22}
\end{minipage}
\begin{minipage}{0.5\hsize}
\begin{center}
\includegraphics[width=1\hsize]{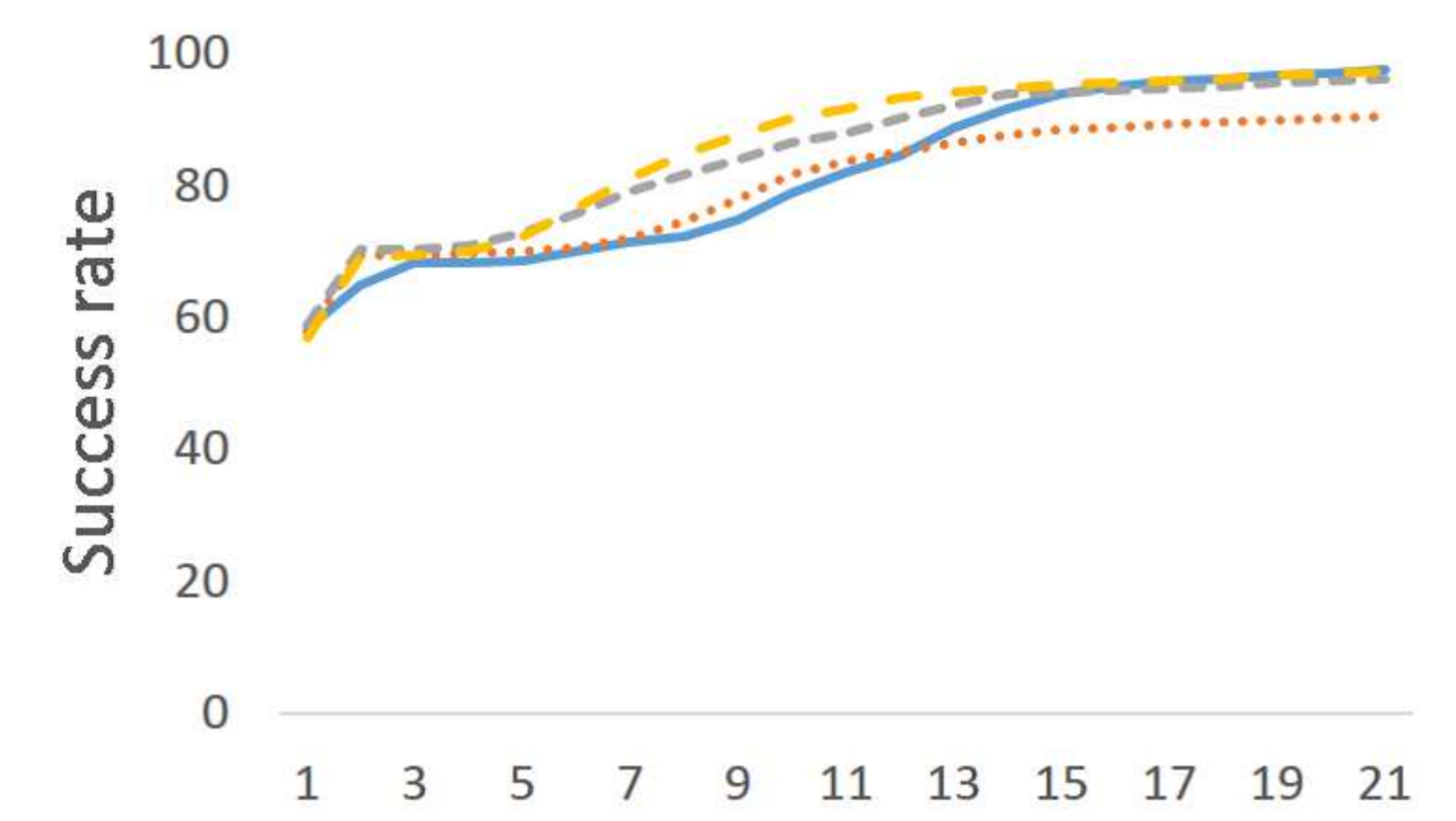}
\end{center}
    \vspace{-14pt}
\subcaption{\textbf{SB}$\wedge$\textbf{RandU}}
\label{fig:21}
\end{minipage}\begin{minipage}{0.5\hsize}
\begin{center}
\includegraphics[width=1\hsize]{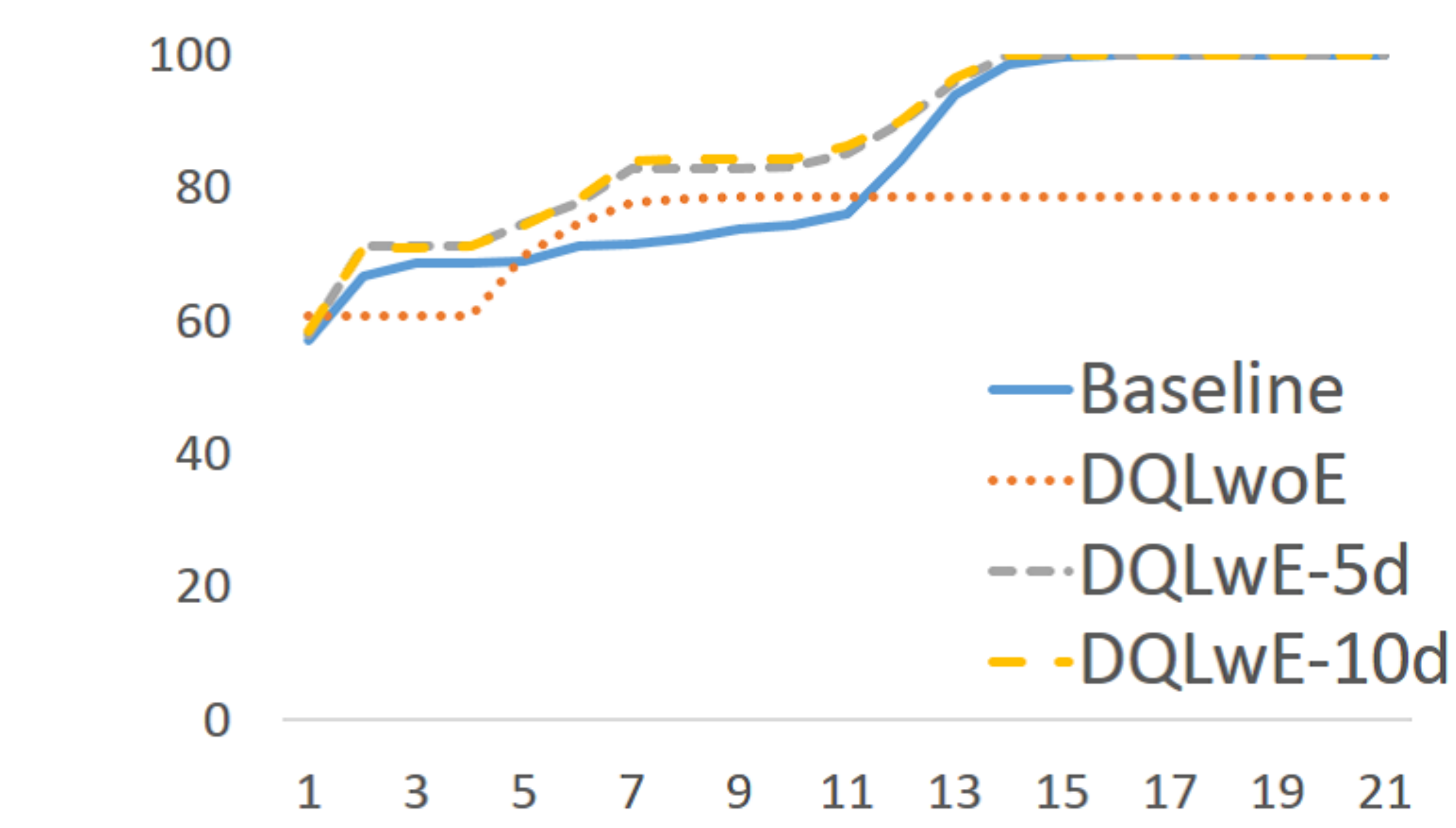}
\end{center}
    \vspace{-14pt}
\subcaption{\textbf{SB}$\wedge$\textbf{RuleU}}
\label{fig:22}
\end{minipage}

\vspace{-10pt}
\caption{Results of each setup. Vertical axis represents the success rate and horizontal axis represents the number of turns.}
\label{fig:sr} 
\vspace{-3pt}
\end{figure}

From the results (Fig.~\ref{fig:sr}), we see that the policies produced by the DRL with logical formula embeddings (DQLwE-5d and DQLwE-10d) performed better than the other policies (Baseline and DQLwoE), where the system has control over the dialog. 
With the exception of the UB cases, DQLwE-5d and DQLwE-10d achieved high success rates in fewer turns than the other policies. 
However, in the UB cases, performances of all policies were more or less the same. 
The main reason is that in the UB cases, the system does not have any beliefs concerning compliance violation (i.e., domain beliefs, as described in Section \ref{sec:framework}). 
If the system does not  have domain beliefs, it cannot perform Open (i.e., condition 2) in $L_{close, i}$ is never satisfied). Therefore, the system cannot control the dialog at all. 

\section{Conclusion and Future Work}\label{sec:conc}
We proposed a method for learning IDS policies and logical formula embeddings in a common framework and found that the combination of the DRL and the logical formula embedding framework performed as effectively or even better than the policies of the hand-crafted baseline. 
In the future, we plan to evaluate our proposed framework in domains in which the system works with users having thousands of beliefs. 
Further, we will evaluate our framework in a more realistic setting where users are real humans and where speech recognition and language understanding errors are included. 

\textbf{Acknowledge:} We gratefully thank Ayako Hoshino, Daniel Andrade, Khurana Jatin, and Vera G\"{o}tzmann for their insightful comments. 

\clearpage
\bibliographystyle{IEEEtran}
\bibliography{mybib}

\end{document}